\documentclass{article}
\usepackage{spconf,amsmath,graphicx,amsfonts, hyperref}
\usepackage{xcolor}
\usepackage{multirow}
\usepackage{booktabs}
\usepackage{makecell}

\newcommand{\rescell}[2]{\makecell{$#1{\scriptstyle \pm#2}$}}

\title{Leveraging Label Correlations in a Multi-label Setting:\\ A Case Study in Emotion}

\name{\makecell{Georgios Chochlakis$^{1, 2}$ \quad Gireesh Mahajan$^3$ \quad Sabyasachee Baruah$^{1, 2}$ \\ Keith Burghardt$^2$ \qquad Kristina Lerman$^2$ \qquad Shrikanth Narayanan$^{1, 2}$}}
\address{\normalsize $^1$ Signal Analysis and Interpretation Lab, University of Southern California, Los Angeles, CA 90089, USA \\ \normalsize $^2$ Information Science Institute, University of Southern California, Marina del Rey, CA 90292, USA \\ \normalsize $^3$ Microsoft Cognitive Services, Redmond, WA 98052, USA}

\begin{document}
\ninept
\maketitle
\begin{abstract}
Detecting emotions expressed in text has become critical to a range of fields. In this work, we investigate ways to exploit label correlations in multi-label emotion recognition models to improve emotion detection.
First, we develop two modeling approaches to the problem in order to capture word associations of the emotion words themselves, by either including the emotions in the input, or by leveraging Masked Language Modeling (MLM). Second, we integrate pairwise constraints of emotion representations as regularization terms alongside the classification loss of the models. We split these terms into two categories, \textit{local} and \textit{global}. The former dynamically change based on the gold labels, while the latter remain static during training.
We demonstrate state-of-the-art performance across Spanish, English, and Arabic in SemEval 2018 Task 1 E-c using monolingual BERT-based models. On top of better performance, we also demonstrate improved robustness. Code is available at \url{https://github.com/gchochla/Demux-MEmo}.\footnote{Funded in part by DARPA under contract HR001121C0168}
\end{abstract}

\vspace{-0.1cm}

\vspace{-0.4cm}
\section{Introduction}
\label{sec:intro}

Emotions are fundamental to human experience. They shape what people pay attention to and how they consume information, what they believe, and how they interact with others~\cite{van2011emotion,dukes2021rise,wahl2019emotions}. Recent advances in deep learning have enabled extraction of emotion signals from language~\cite{calvo2015oxford, alhuzali2021spanemo, yu2018improving, baziotis2018ntua}, thereby facilitating emotion recognition from text at scale~\cite{guo2022emotion}. 
Despite these successes, the need for more accurate, robust and fair emotion recognition models remains.

We focus on inferring categorical emotions from text, like those in Plutchik's wheel of emotions \cite{plutchik1980general}, a theory of emotion that portrays discrete emotions on a 2D space. Early textual emotion recognition models have used hand-crafted features and emotion lexicons \cite{stone1966general, strapparava2004wordnet, pennebaker2001linguistic}. Modern deep learning models achieve better performance \cite{alhuzali2021spanemo, ying2019improving, baziotis2018ntua} but have to contend with the sampling bias of the data, and the subjective nature of the annotations during training.

\begin{figure}[t]
\centerline{
 \includegraphics[width=1.0\columnwidth]{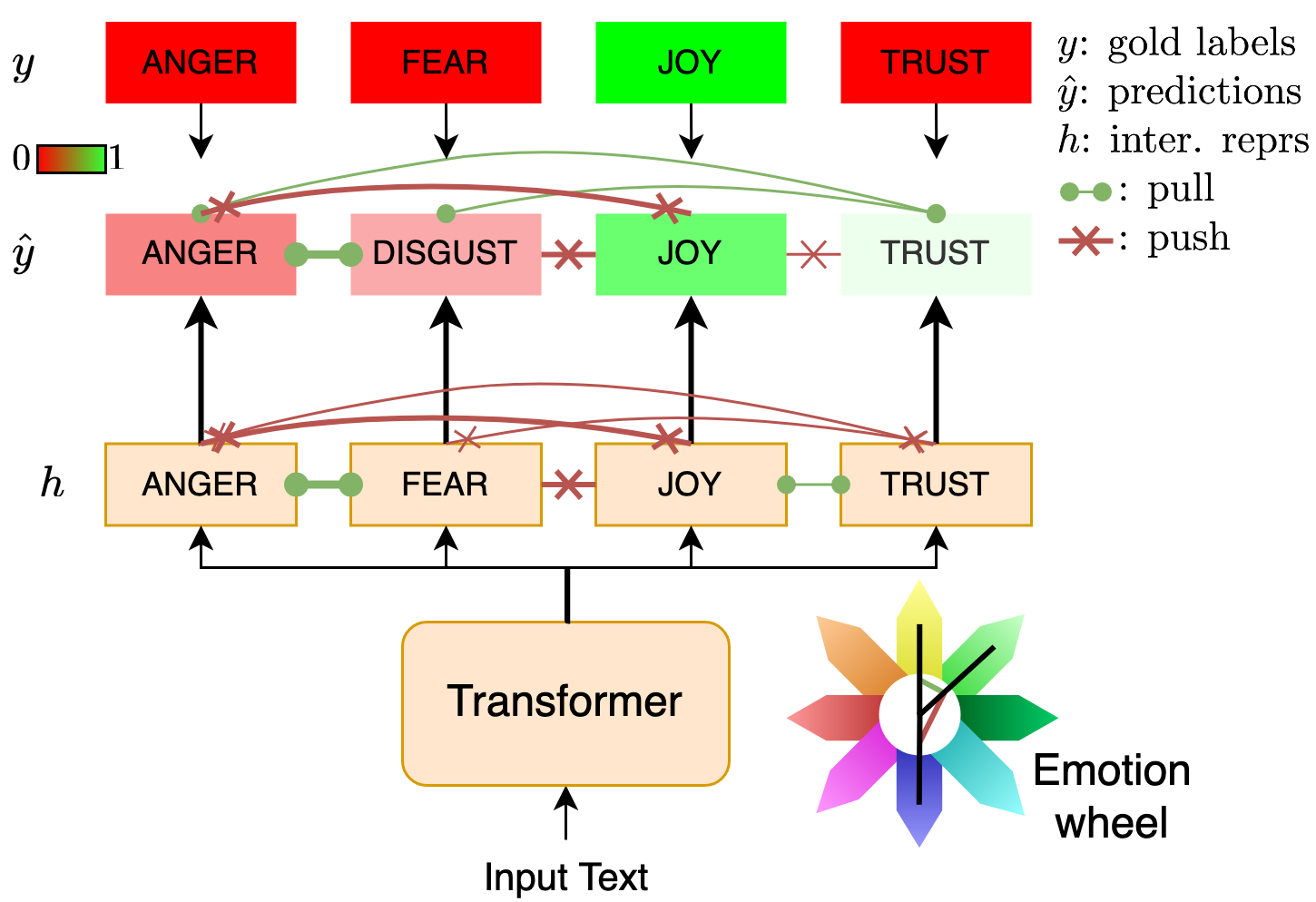}}
 \caption{We induce label-correlation awareness by pulling together or pushing apart representations of pairs of emotions. This can be achieved at the level of intermediate representations $h$, and at the level of predictions $\hat{y}$. Prediction pairs can be regularized using the labels $y$, by pulling emotions with the same gold labels together, otherwise pushing them apart. Representation pairs can be regularized in the same way, but we can also use prior relationships between them, which can conflict with current labels. Each pair's regularization term can be modulated by the strength of the relationship of the pair.}
 \label{fig:thumbnail}
\end{figure}

Furthermore, we concentrate on the more general setting of multi-label emotion recognition, meaning none, one, or multiple emotions can be present per example. This multi-label setting is more realistic and computationally intricate because the model's predictions correlate with each other.

We study how we can integrate these relationships in Transformer-based \cite{vaswani2017attention} models beyond the regular training. First, we consider two different modeling approaches to leverage word-level associations in the language model (LM). 
In the first one, we leverage global attention and multiple input sequences. Borrowing from \cite{alhuzali2021spanemo}, we directly use the emotions in the input by prepending them to the input text. Thereafter, their contextual embeddings are used for the final classification. In this manner, word-level associations between the emotions arise due to the attention mechanism. We refer to this modeling approach as \verb+Demux+ (short for ``Demultiplexer''), because the emotions, provided in the input, can be thought of as \textit{selectors} of features from the input text sequence we want to classify.

The second formulation is based on the Masked Language Modeling (MLM) pretraining task. We create a more natural prompt, a declarative statement about the emotional content of the input sequence, but with the actual emotion(s) replaced by a \verb+[MASK]+ token. \verb+[MASK]+'s contextual embedding is then fed to the pretrained MLM prediction head or a new classifier to perform emotion recognition. Word-level associations are captured by \verb+[MASK]+ and the MLM head. We refer to this approach as \verb+MEmo+ (\textbf{M}asked \textbf{EMO}tion).

The second important question we pose is whether the classification loss is adequate to instill in the models the relationships between emotions (Figure \ref{fig:thumbnail}). We do so by integrating several regularization losses into our aforementioned models. We split these into two main categories, \textit{global} and \textit{local} terms. The former consist of static ``contraints'' based on prior relationships between emotions. The latter are dynamic, as they take into consideration the gold labels of each example. For both, we limit our experiments to pairs of emotions. Since different pairs of emotions relate more or less strongly, we also optionally apply a weighting scheme per pair. 

We show roughly equivalent performance from \verb+Demux+ and \verb+MEmo+. We also observe a trend for \textit{not} pushing emotion representations apart performing favorably. We improve the state of the art for SemEval~2018~Task~1 E-c \cite{mohammad2018semeval}, as well as the robustness of performance across experiments and settings, further emphasizing the utility of our approach. Our contributions can be summarized as:

\begin{itemize}
    \item We demonstrate state-of-the-art performance for SemEval~2018 Task~1 E-c, a widely used, realistic benchmark, using monolingual BERT-based models.
    \item We investigate two different modeling approaches, \verb+MEmo+, based on MLM, and \verb+Demux+, relying on label embeddings, to integrate label correlations through word-level associations.
    \item We study how regularization losses on pairs of emotions can facilitate robustness and label-correlation awareness.
\end{itemize}

\section{Related Work}
\label{sec:related}

Earlier works relied on emotion lexicons and Bag-of-Words (BoW) algorithms. The General Inquirer \cite{stone1966general}, one of the earliest efforts at automated affective analyses on text, is a general-purpose lexicon with some affect-related quantities. More modern efforts include LIWC \cite{pennebaker2001linguistic}, which is widely used, e.g. in Computational Social Science \cite{mehl2017natural}. For more sophisticated BoW methods, DDR \cite{garten2018dictionaries} extends lexicon-based methods from counting to word similarities.
The most recent and successful methods deploy deep learning. Initially, single-label predictions were transformed into a multi-label output with a probability threshold \cite{he2018joint}. LSTMs \cite{hochreiter1997long} have been widely used for the task, e.g., the usage of DeepMoji features \cite{felbo2017using} by SeerNet \cite{duppada2018seernet}, and NTUA-SLP \cite{baziotis2018ntua} for SemEval 2018 Task 1 \cite{mohammad2018semeval}. It is interesting that these methods also leveraged features from affective lexicons. Attention and transfer learning is also studied in~\cite{yu2018improving}.
More recently, Transformers~\cite{vaswani2017attention} have dominated the field. They have been coupled with Convolutional Neural Networks~\cite{ying2019improving} or Graph Neural Networks~\cite{xu2020emograph}. \verb+SpanEmo+~\cite{alhuzali2021spanemo} prompts a BERT model by enumerating all emotions in its input and uses a Label-correlation aware (LCA) loss to facilitate emotion interplay.

In addition, language prompting has garnered attention because it allows language transformers to perform different tasks under the same framework, as well as zero and few-shot inference~\cite{brown2020language, liu2021pre, reed2022generalist}.

\section{Methodology}

Label correlations can be considered at the \textit{local} and the \textit{global} level. For global supervision, we rely on prior information about the label relations, such as an empirical covariance matrix, or theoretically-inspired e.g.,  Plutchik's wheel of emotions \cite{plutchik1980general}. We can leverage those to align intermediate representations $h$ of a model. On the local level, however, annotations of specific examples $y$ can indicate conflicting relationships to our priors, and predictions of the network $\hat{y}$ can also be guided. In terms of modeling, we leverage the relationships between emotion words to facilitate label correlations, either by including them in the input or by using a \verb+[MASK]+ token.

\label{sec:method}

Given $n$ emotions, let $E = \{e_i: i\in[n]\}$ be the set of emotions and $c: [n]^2 \rightarrow [0, 1]$ a measure of correlation between them, like the correlation of the train set, $\rho$, or the cosine in Plutchik's wheel \cite{plutchik1980general}, $\theta$, both projected to $[0, 1]$ from $[-1, 1]$. Let $\mathcal{L}$ denote loss functions.

\subsection{Correlation-aware Regularization}

In an effort to enhance the model's correlation awareness between emotions, we introduce additional terms to the classification loss. They can be categorized into two groups, \textit{global} and \textit{local}.

\vspace{-0.2cm}
\subsubsection{Global}

Global supervision refers to any loss that constrains the distance of pairs of emotions based on prior relationships between them, and is, therefore, fixed during training and applied to intermediate emotion representations. Given representations per emotion $h$:
\begin{equation}
    \mathcal{L}_G(h; c) = \frac{1}{n^2 - n} \sum_{(i, j) \in [n]^2}^{i\ne j} (\text{cossim}(h_i, h_j) - c_{i, j})^2
\end{equation}

\vspace{-0.5cm}
\subsubsection{Local}

Local supervision takes into account the gold labels for each example in the formulation of the loss. Therefore, the emotions are split into two groups, the present and the absent emotions in $y$, $\mathcal{P}$ and $\mathcal{N}$ respectively. We consider three types of local losses. First, those that dictate \textbf{inter}-group relationships between $\mathcal{P}$ and $\mathcal{N}$, where we try to increase the distance of representations of pairs of emotions. The hypothesis here is that the presence of one emotion informs us what emotions are likely to be absent, and vice versa. Then, we have those that dictate \textbf{intra}-group relationships within $\mathcal{P}$ and within $\mathcal{N}$, where we try to decrease the distance of each pair. The assumption here is that a present emotion informs us what other emotions are likely to be present, and similarly for absent emotions. Finally, we have their combination, where we simultaneously push apart inter-group pairs and pull together intra-group pairs. Our formulation is:
\begin{equation}
    \begin{split}
        \mathcal{L}_{L, \text{inter}}(y, r; S, f) = & \frac{1}{|\mathcal{N}| |\mathcal{P}|} \sum_{i\in \mathcal{N},\ j\in \mathcal{P}} f_{i, j} \cdot S(r_i, r_j) \\
        \mathcal{L}_{L, \text{intra}}(y, r; D, f^\prime) =
            & \frac{1}{2} \left[ \frac{2}{|\mathcal{N}|^2 - |\mathcal{N}|} \sum_{(i, j) \in \mathcal{N}^2}^{i > j} f^\prime_{i, j} \cdot D(r_i, r_j) \quad \right. \\
            + & \left. \frac{2}{|\mathcal{P}|^2 - |\mathcal{P}|} \sum_{(i, j) \in \mathcal{P}^2}^{i > j} f^\prime_{i, j} \cdot D(-r_i, - r_j) \right] \\
        \mathcal{L}_{L, \text{both}}(y, r; S, D, f, f^\prime&) = \frac{\mathcal{L}_{L, \text{inter}}(y, r; S, f) + \mathcal{L}_{L, \text{intra}}(y, r; D, f^\prime)}{2}, \\
    \end{split}
    \label{eq:locals}
\end{equation}
where $S$ quantifies ``similarity'' and $D$ ``distance'' between some representations of the network $r$, and $f$, $f^\prime$ are weights for each pair of emotions. $S$ and $D$ do not have to strictly adhere to similarity/distance properties, but decreasing $S$ should force the emotion representations to diverge, while $D$ should pull them closer. The denominators average the terms to keep the magnitude of the different losses similar. Borrowing from the LCA loss in \cite{alhuzali2021spanemo}, when we set $f = f^\prime = 1$, $r$ as the output probabilities of the model $\hat{y}$ and
\begin{equation} \label{eq:eD}
    S(\hat{y}_i, \hat{y}_j) = e^{\hat{y}_i - \hat{y}_j}, \quad D(\hat{y}_i, \hat{y}_j) = e^{\hat{y}_i + \hat{y}_j},
\end{equation}
we retrieve $\mathcal{L}_{L, \text{inter}} = \mathcal{L}_{LCA}$. Note that this formulation of $D$ requires the negative signs in $\mathcal{L}_{L, \text{intra}}$ because present emotions would be pushed to $0$ otherwise. Another alternative is setting $r$ as some intermediate representations $h$ and
\begin{equation} \label{eq:cossim}
    S(h_i, h_j) = \text{cossim}(h_i, h_j), \quad D(h_i, h_j) = -\text{cossim}(h_i, h_j).
\end{equation}
Note that for this formulation $D(-r_i, - r_j) = D(r_i, r_j)$.

We further augment our local losses by making $f$ and $f^\prime$ functions of $c$. We pick $f$ to be decreasing, and $f^\prime$ to be increasing, so that each pair is weighted proportionately to the probability it would end up in that group state. For example, the weight of a frequently co-occurring pair becomes small when they appear with different gold labels in $\mathcal{L}_{L, \text{inter}}$. We fix $f_{i, j} = 1 - c_{i, j}$ and $f^\prime_{i, j} = c_{i, j}$.

The final loss we use is
\begin{equation} \label{eq:loss}
    \mathcal{L} = (1 - \alpha)\mathcal{L}_{BCE} + \alpha \mathcal{L}_L + \beta \mathcal{L}_G
\end{equation}

\vspace{-0.7cm}
\subsection{Models}

\subsubsection{Demux}

Let $x$ be an input text sequence. Inspired by \cite{alhuzali2021spanemo}, we use emotions $E$ to construct $x^\prime=\text{``}e_1, e_2, \dots, \text{or}\ e_n?\text{''}$, and use tokenizer $T$:
\begin{equation}
\begin{split}
    \tilde{x} = T(x^\prime, x) = (&[CLS], t_{1, 1}, \dots, t_{1, N_1}, \dots,
    \\ & t_{n, 1}, \dots, t_{n, N_n}, [SEP], x_1, \dots, x_l),
\end{split}
\end{equation}
where $x_i$ are the tokens from $x$, $t_{i, j}$ is the $j$-th subtoken of $e_i$, and $T$ typically will combine the given sequences with \verb+[SEP]+, and prepend \verb+[CLS]+ to the entire sequence. We pass $\tilde{x}$ through LM $L$ to finally get $\hat{x} = L(\tilde{x})$, where $\hat{x}$ contains one output embedding corresponding to each input token. We denote the output embedding corresponding to $t_{i,j}$ as $\hat{t}_{i, j}\in\mathbb{R}^d$, where $d$ is the feature dimension of $L$. Thereafter, we aggregate the outputs of the subtokens corresponding to each class by averaging them, and get the final probabilities $p(e_i|x)$ using a 2-layer neural network, $\text{NN}: \mathbb{R}^d \rightarrow \mathbb{R}$, followed by the sigmoid function $\sigma$:
\begin{equation}
    \forall i \in [n],\quad p(e_i|x) = \sigma(\text{NN}(\frac{\sum_{j=1}^{N_i}\hat{t}_{i, j}}{N_i}))).
\end{equation}

\begin{table}[!ht]
    \centering
    \begin{tabular}{lc@{\hspace{0.8\tabcolsep}}c@{\hspace{0.95\tabcolsep}}cc@{\hspace{0.95\tabcolsep}}c@{\hspace{1\tabcolsep}}c@{\hspace{1\tabcolsep}}c}
        &\multicolumn{3}{c}{$\mathcal{L}_L$} && \multicolumn{3}{c}{En SemEval 2018 Task 1 E-c} \\
        \cmidrule{2-4} \cmidrule(lr){6-8}
        & Group & $f, f^\prime$ & $S, D$ & $\mathcal{L}_G$ & Mic-F1 & Mac-F1 & JS \\
        \toprule
        
        \parbox[t]{2mm}{\multirow{22}{*}{\rotatebox[origin=c]{90}{Demux}}} & - & - & - & - & \rescell{72.0}{0.3} & \rescell{57.1}{1.6} & \rescell{60.9}{0.3} \\
        
        \cmidrule{2-8}
        
        &- & - & - & $\rho$ & \rescell{72.0}{0.3} & \rescell{57.1}{1.6} & \rescell{60.9}{0.3} \\
        &- & - & - & $\theta$ & \rescell{72.0}{0.4} & \rescell{56.7}{1.5} & \rescell{60.9}{0.4} \\
        
        \cmidrule{2-8}
        
        &inter & & & & \rescell{72.1}{0.3} & \rescell{57.8}{1.5} & \rescell{60.9}{0.3} \\
        &intra & 1 & $e^y$ & - & \rescell{\mathbf{72.2}}{0.3} & \rescell{\mathbf{58.2}}{1.4} & \rescell{61.0}{0.3} \\
        &both & & & & \rescell{72.0}{0.3} & \rescell{57.5}{1.2} & \rescell{60.9}{0.6} \\
        
        \cmidrule{2-8}
        
        &inter & & & & \rescell{72.1}{0.4} & \rescell{57.6}{1.0}  & \rescell{60.9}{0.3}  \\
        &intra & $\rho$ & $e^y$ & - & \rescell{\mathbf{72.3}}{0.4} & \rescell{\mathbf{58.1}}{1.5} & \rescell{\mathbf{61.2}}{0.4} \\
        &both & & & & \rescell{72.1}{0.3} & \rescell{56.4}{1.6} & \rescell{61.0}{0.3} \\
        
        \cmidrule{2-8}
        
        &inter & & & & \rescell{72.0}{0.4} & \rescell{57.8}{0.6} & \rescell{60.9}{0.4} \\
        &intra & $\theta$ & $e^y$ & - & \rescell{\mathbf{72.2}}{0.5} & \rescell{57.9}{1.3} & \rescell{\mathbf{61.1}}{0.4} \\
        &both & & & & \rescell{\mathbf{72.3}}{0.4} & \rescell{\mathbf{58.1}}{1.1} & \rescell{\mathbf{61.3}}{0.4} \\
        
        \cmidrule{2-8}
        
        & & 1 & & & \rescell{71.9}{0.3} & \rescell{56.7}{1.3} & \rescell{60.8}{0.3} \\
        &both & $\rho$ & $\cos$ & - & \rescell{71.7}{0.3} & \rescell{57.1}{1.5} & \rescell{60.5}{0.3} \\
        & & $\theta$ & & & \rescell{71.8}{0.3} & \rescell{57.1}{1.8} & \rescell{60.6}{0.4} \\
        
        \cmidrule{2-8}
        
        &intra & 1 & & & \rescell{72.2}{0.4} & \rescell{\mathbf{58.3}}{1.3} & \rescell{61.0}{0.3} \\
        &intra & $\rho$ & $e^y$ & $\rho$ & \rescell{72.2}{0.3} & \rescell{57.8}{1.6} & \rescell{\mathbf{61.3}}{0.4} \\
        &both & $\theta$ & & & \rescell{72.0}{0.4} & \rescell{58.0}{1.4} & \rescell{60.9}{0.5} \\

        \toprule \toprule

        \parbox[t]{2mm}{\multirow{12}{*}{\rotatebox[origin=c]{90}{MEmo}}} & - & - & - & - & \rescell{71.1}{0.4} & \rescell{55.9}{1.6} & \rescell{59.9}{0.5} \\

        \cmidrule{2-8}
        
        &inter & & & & \rescell{71.6}{0.3} & \rescell{56.9}{1.5} & \rescell{60.3}{0.4} \\
        &intra & 1 & $e^y$ & - & \rescell{\mathbf{72.1}}{0.3} & \rescell{\mathbf{57.7}}{1.8} & \rescell{\mathbf{61.0}}{0.4} \\
        &both & & & & \rescell{71.8}{0.3} & \rescell{57.1}{1.3} & \rescell{60.6}{0.5} \\
        
        \cmidrule{2-8}
        
        &inter & & & & \rescell{71.5}{0.4} & \rescell{57.0}{1.3} & \rescell{60.2}{0.5} \\
        &intra & $\rho$ & $e^y$ & - & \rescell{71.8}{0.4} & \rescell{57.1}{1.4} & \rescell{60.6}{0.5} \\
        &both & & & & \rescell{71.6}{0.4} & \rescell{56.5}{1.8} & \rescell{60.3}{0.4} \\
        
        \cmidrule{2-8}
        
        &inter & & & & \rescell{71.5}{0.3} & \rescell{\mathbf{57.7}}{1.1} & \rescell{60.2}{0.3} \\
        &intra & $\theta$ & $e^y$ & - & \rescell{71.6}{0.5} & \rescell{56.7}{1.3} & \rescell{60.4}{0.4} \\
        &both & & & & \rescell{71.6}{0.4} & \rescell{57.1}{1.2} & \rescell{60.3}{0.5} \\
        
    \end{tabular}
    \caption{Performance on English SemEval 2018 Task 1 E-c dev set. \textit{intra}, \textit{inter} and \textit{both} refer to $\mathcal{L}_{L, \text{inter}}$, $\mathcal{L}_{L, \text{intra}}$ and $\mathcal{L}_{L, \text{both}}$ respectively (Eq. \ref{eq:locals}), $\rho$, $\theta$, and 1 to setting $c$ as the empirical correlation, the wheel angles and (for $\mathcal{L}_L$) keeping weights of pairs constant respectively, and $e^y$ and $\cos$ refer to Eq. \ref{eq:eD} and Eq. \ref{eq:cossim} respectively.}
    \label{tab:analysis}
\end{table}

\vspace{-0.7cm}
\subsubsection{MEmo}

Let $x$ be an input text sequence. We prepend ``emotion \verb+[MASK]+ in tweet \textvisiblespace'' to $x$ to create our input, $\tilde{x}$. We then use $L$ with its tokenizer $T$ to get the output $\hat{x} = L(T(\tilde{x}))$, where $\hat{x}$ contains one output embedding corresponding to each input token. From $\hat{x}$, we extract the embedding that corresponds to \verb+[MASK]+, $h_M$. We can use two methods to get the final predictions, the pretrained MLM head, or build a new classifier. When using the MLM head, for each emotion, we use its corresponding logit and apply the sigmoid. For the latter, to get the probability $p(e_i|x)$, we pass $h_M$ through a 2-layer neural network, $\text{NN}: \mathbb{R}^d \rightarrow \mathbb{R}^{|E|}$ and then apply the sigmoid
\begin{equation}
    \forall i \in [n],\quad p(e_i|x) = \sigma(\text{NN}(h_M)_i).
\end{equation}

We find there is no meaningful way to implement hidden representations per emotion for this model, hence losses that depend on the cosine similarity are not integrated.

\section{Experiments} \label{sec:exp}

\begin{table}[t]
    \centering
    \begin{tabular}{l@{\hspace{0.85\tabcolsep}}c@{\hspace{0.85\tabcolsep}}cccc}
        &&& \textbf{Mic-F1} & \textbf{Mac-F1} & \textbf{JS} \\
        \cmidrule(lr){4-6}
        Model & $\mathcal{L}_L$ & $\mathcal{L}_G$ & \multicolumn{3}{c}{En SemEval 2018 Task 1 E-c} \\

        \cmidrule(l){1-3} \cmidrule(lr){4-6}
        
        \verb+SpanEmo+$^\dagger$ & inter 1 $e^y$&-&
            \rescell{69.9}{0.8} & \rescell{52.8}{1.4} & \rescell{57.4}{1.0} \\

        \verb+Demux+ & - & - &
            \rescell{71.7}{0.2} & \rescell{54.9}{0.4} & \rescell{60.1}{0.2} \\
        \verb+Demux+ & intra 1 $e^y$ & - &
            \rescell{72.2}{0.2} & \rescell{\mathbf{58.1}}{0.5} & \rescell{60.8}{0.2} \\
        \verb+Demux+ & intra $\theta\ e^y$ & - &
            \rescell{72.2}{0.2} & \rescell{57.1}{0.3} & \rescell{60.7}{0.2} \\
        \verb+Demux+$^*$ & intra $\rho\ e^y$ & - &
            \rescell{72.2}{0.2} & \rescell{57.6}{0.3} & \rescell{60.8}{0.2} \\

        \verb+MEmo+$^*$ & intra 1 $e^y$ & - & \rescell{\mathbf{72.3}}{0.2} & \rescell{56.1}{0.5} & \rescell{\mathbf{61.1}}{0.2} \\

        \cmidrule(lr){4-6}

        &&& \multicolumn{3}{c}{Es SemEval 2018 Task 1 E-c} \\
        \cmidrule(lr){4-6}
        
        \verb+SpanEmo+$^\dagger$ & inter $e^y$ & - &
            \rescell{59.5}{1.3} & \rescell{46.8}{3.0} & \rescell{47.7}{1.7} \\

        \verb+Demux+ & - & - &
            \rescell{62.7}{0.4} & \rescell{54.1}{0.4} & \rescell{54.2}{0.4} \\
        \verb+Demux+ & intra 1 $e^y$ & - &
            \rescell{63.2}{0.4} & \rescell{\mathbf{54.2}}{0.5} & \rescell{54.7}{0.3} \\
        \verb+Demux+$^*$ & intra $\rho\ e^y$ & - &
            \rescell{63.0}{0.2} & \rescell{\mathbf{54.2}}{0.3} & \rescell{54.4}{0.3} \\
        
        \verb+MEmo+$^*$ & intra 1 $e^y$ & - & \rescell{\mathbf{63.3}}{0.4} & \rescell{51.7}{0.6} & \rescell{\mathbf{55.6}}{0.4} \\

        \cmidrule(lr){4-6}

        &&& \multicolumn{3}{c}{Ar SemEval 2018 Task 1 E-c} \\
        \cmidrule(lr){4-6}
        \verb+SpanEmo+$^\dagger$ & inter 1 $e^y$ & - &
            \rescell{65.8}{0.9} & \rescell{50.5}{1.9} & \rescell{53.5}{1.1} \\

        \verb+Demux+ & intra 1 $e^y$ & $\rho$ &
            \rescell{67.6}{0.5}  & \rescell{53.1}{0.8} & \rescell{56.0}{0.5} \\
        \verb+Demux+$^*$ & intra $\rho\ e^y$ & - &
            \rescell{67.5}{0.5} & \rescell{\mathbf{53.8}}{0.8} & \rescell{56.0}{0.6} \\
        
        \verb+MEmo+$^*$ & intra 1 $e^y$ & - & \rescell{\mathbf{67.9}}{0.4} & \rescell{51.6}{0.5} & \rescell{\mathbf{56.7}}{0.5} \\
       
    \end{tabular}
    \caption{Performance on SemEval 2018 Task 1 E-c test set. \textit{intra} and \textit{inter} refer to $\mathcal{L}_{L, \text{inter}}$ and $\mathcal{L}_{L, \text{intra}}$ respectively (Eq. \ref{eq:locals}), $\rho$, $\theta$, and 1 to setting $c$ as the empirical correlation, the wheel angles and (for $\mathcal{L}_L$) keeping weights of pairs constant respectively, and $e^y$ refers to Eq. \ref{eq:eD}. $^\dagger$: reproduced from authors' code. $^*$: main result.}
    \label{tab:sota}
\end{table}

\subsection{Datasets}

We use the publicly available SemEval 2018 Task 1 E-c \cite{mohammad2018semeval}, which contains tweets annotated for 11 emotions in a multi-label setting, namely \textit{anger, anticipation, disgust, fear, joy, love, optimism, pessimism, sadness, surprise,} and \textit{trust}, in three languages, \textit{English, Arabic,} and \textit{Spanish}.
The English subset contains 6838 training, 886 development and 3259 testing tweets, the Arabic subset contains 2278 training, 585 development and 1518 testing tweets, and the Spanish subset 3561 training, 679 development and 2854 testing tweets. We use the Jaccard Score (JS), macro F1 and micro F1 for all our evaluations.

\subsection{Implementation Details}

We use Python (v3.7.4), PyTorch (v1.11.0) and the corresponding implementations of Transformer models from the Hugging Face \textit{transformers} library (v4.19.2). We use up to three NVIDIA GeForce GTX 1080 Ti and one NVIDIA GeForceRTX 2070, but always one GPU per model.

We use BERT for English \cite{devlin2018bert} and Arabic \cite{safaya-etal-2020-kuisail}, and BETO \cite{CaneteCFP2020} for Spanish, as in \cite{alhuzali2021spanemo}. In terms of hyperparameters, we retain the hyperparameters used in \cite{alhuzali2021spanemo}, such as the learning rate and its schedule, batch size, and $\alpha$ (Eq. \ref{eq:loss}). However, for the text preprocessor, we further remove the special tags introduced by the preprocessing library, \textit{ekphrasis}\footnote{\url{https://github.com/cbaziotis/ekphrasis}}. We found $0.1$ and $0.5$ to work equally well for $\beta$ (Eq. \ref{eq:loss}), so we set $\beta = 0.1$ when we use the global loss. For early stopping, we reduce the patience to 5 epochs, and use JS instead as the evaluation metric. Moreover, instead of directly using the model after early stopping, we retrain it on both the train and dev sets, picking the number of epochs based on the performance distribution of the model across multiple runs.
Test and dev performance are reported after 10 runs.

\subsection{Local and Global relations}

We run comprehensive experiments to determine the better alternative w.r.t. regularization losses and present our results in Table \ref{tab:analysis}. In general, for \verb+Demux+, we see minimal changes in performance from configuration to configuration. As general trends, we observe that reinforcing intra-group relations works better compared to inter-group ones, or even their combination. This implies that creating opposing forces between groups of emotions can hurt performance, perhaps because the ability to express more nuance with multiple different emotions is hindered. The global regularization and our weighting scheme seem to have little to no effect, and the local regularization based on the cosine similarity seems to be too strict, noticeably degrading performance even compared to the baseline.

For \verb+MEmo+, we see better performance with the new classifier instead of the MLM head, thus we report the former's numbers. We observe reduced macro F1 but similar micro F1 and JS to \verb+Demux+. We see clearer improvement by the regularization terms, suggesting this alternative does not capture word-level associations as well as \verb+Demux+ does, and similar trends with intra vs. inter-group terms.

\subsection{State-of-the-art}

In this section, we compare only with the previous state of the art, \verb+SpanEmo+~\cite{alhuzali2021spanemo}, due to space limitations. Results can be seen in Table~\ref{tab:sota}. Given that the original code has been open-sourced~\footnote{\url{https://github.com/gchochla/SpanEmo}}, we also reproduce their results to derive averages and deviations. 

First, comparing \cite{alhuzali2021spanemo} with \verb+Demux+, we see a significant increase in performance no matter the latter's configuration. In fact, ranges never overlap and \textit{all} our improvements are \textit{statistically significant} based on t-tests with Bonferonni correction ($p<1.5\cdot10^{-3}$). Looking at micro-F1 and macro-F1, which the model was not explicitly tuned to perform well at, we see $3-6\%$ relative improvements for the former, and $6-15\%$ for the latter. For JS, relative improvements are around $5-14\%$. Note that, for consistency, we report test performance of ``intra $\rho\ e^y$'' with $\beta=0$ across all languages to facilitate reporting in future work. \verb+MEmo+ also outperforms the previous state of the art with statistically significant improvements ($p<4\cdot10^{-4}$) but for macro F1 in Arabic ($p\simeq0.112$). Micro F1, macro F1 and JS improve by $3-6\%$, $2-10\%$, and $6-16\%$ over \cite{alhuzali2021spanemo} respectively. Comparing our two models, we consistently see roughly equivalent micro F1, favorable JS from \verb+MEmo+ but a decrease in macro F1.

Compared to the dev set, here we observe improved robustness of the model utilizing some local loss as opposed to the simple baseline, especially in English. This suggests that, while the ``potential'' (i.e., the dev set performance) of the different models is similar, generalization is aided by including local and global losses in the model's supervision. Moreover, as is evident from the standard deviations, our pipeline achieves more robust results \textit{in all configurations}. Performance on individual emotions is shown in our repo.






\section{Conclusion}
\label{sec:conc}

In this work, we examine techniques to induce correlation-awareness in models of emotion recognition in a multi-label setting. We investigated a variety of configurations of global and local regularization losses constraining the behavior of pairs of emotions. We also experimented with architectures and formulations that leverage word associations between the emotion words either in the input, by explicitly including the emotions, or the output, using the MLM pretraining task of the model. We also demonstrate that better configuring the training regime can improve performance and robustness.

We achieve state-of-the-art results with monolingual BERT-based models, with improvements that are statistically significant, and indeed improve substantially over existing work. We also demonstrate that the examined regularization losses can improve the robustness of the resulting model compared to a model solely guided by a classification loss. Finally, we provide guidelines, and open-source our code, in an effort to guide other researchers when extending or repurposing our work.

Our work also demonstrates that cosine-similarity-based losses cannot really improve the models. Future work should try and address these issues, as, intuitively, these losses would be expected to work favorably based on the fact that they explicitly model the relationship between the emotions, whereas the difference of the predictions can be optimized for each emotion separately, and interactions only arise because of the contextual embeddings in earlier layers. Moreover, weighting each term also did not help. Future work could investigate non-linear functions of $c$, general ways to deal with the change of magnitude in the regularization loss, and perhaps estimating the correlations by collating multiple sources, e.g., using a mixed-effects model.



\bibliographystyle{acm}
\bibliography{refs}

\end{document}